\documentclass[journal]{IEEEtran}
\IEEEoverridecommandlockouts
\usepackage[utf8]{inputenc} 
\usepackage[T1]{fontenc}    
\usepackage{hyperref}       
\usepackage{url}            
\usepackage{booktabs}       
\usepackage{amsfonts}       
\usepackage{nicefrac}       
\usepackage{microtype}      

\usepackage{graphicx}
\usepackage{subfigure} 

\usepackage{amssymb}
 \usepackage{amsthm}
 \usepackage{amsmath}
 \usepackage{hyperref}
\usepackage{bm}
\usepackage[switch]{lineno}
\ifCLASSINFOpdf
\else
\fi
\begin{document}

%
\title{Money on the Table: Statistical information ignored by softmax can improve classifier accuracy}
%
%
%

\author{Charles B. Delahunt,~\IEEEmembership{Member,~IEEE,}
        Courosh Mehanian, 
        and~J. Nathan Kutz 
\thanks{C.B. Delahunt and J.N. Kutz are with the Dept of Applied Mathematics, University of Washington, Seattle, WA. e-mail: \{delahunt, kutz\}@uw.edu}
\thanks{C.B. Delahunt and C. Mehanian are with Global Good Research, Bellevue, WA. e-mail: \{cdelahunt, cmehanian\}@intven.com}
}

\maketitle

\begin{abstract}
Softmax is a standard final layer used in Neural Nets (NNs) to summarize information encoded in the trained NN and return a prediction.
However, softmax leverages only a subset of the class-specific structure encoded in the trained model and ignores potentially valuable  information:
During training, models encode an array $\bm{D}$ of class response distributions, where $\bm{D_{ij}}$ is the distribution of the $\bm{j^{th}}$ pre-softmax readout neuron's responses to the $\bm{i^{th}}$ class.
Given a test sample, softmax implicitly uses only the row of this array $\bm{D}$ that corresponds to the   sample's true class.
Leveraging more of this array $\bm{D}$ can improve classifier accuracy, because the likelihoods of two competing classes  can be encoded in other rows of $\bm{D}$.
To explore this potential resource, we develop a hybrid classifier (Softmax-Pooling Hybrid, $\bm{SPH}$) that uses softmax on high-scoring samples, but on low-scoring samples uses a log-likelihood method that pools the information from the full array $\bm{D}$.
We apply $\bm{SPH}$ to models trained on a vectorized MNIST dataset to varying levels of accuracy.
$\bm{SPH}$ replaces only the final softmax layer in the trained NN, at test time only (all training is the same as for softmax).
Because the pooling classifier performs  better than softmax on low-scoring samples, $\bm{SPH}$ reduces test set error by 6\% to 23\%, using the exact same trained model, whatever the baseline softmax accuracy.
This  reduction in error reflects hidden capacity of the trained NN that is  unused by softmax.
\end{abstract}


%
\IEEEpeerreviewmaketitle

\section{Introduction}

The softmax  is a standard final layer in classifiers such as Neural Nets (NNs) and Support Vector Machines (SVMs). 
It interprets the outputs of the NN to deliver an estimated class for test samples. 
This is a logical choice because NNs are typically trained on a loss function with an embedded softmax.
Softmax also scales to large number of classes, and returns values that can be interpreted as probabilities.
In a highly-trained NN, softmax will ideally return a value $\approx$1 for the predicted (and hopefully correct) class and $\approx$0 for all other classes. 

However, not all situations allow for such a highly-trained NN. 
For example, there may be insufficient training data available, as is often the case for medical, scientific, or field-collected datasets where data collection is difficult or costly \cite{koller2018}.
This can lead to a mis-match of problem complexity and training data: 
A certain model complexity  is required to capture the problem well enough to meet use-case performance specs, but  insufficient training data exists to properly fit the complex model.
One standard method in this situation is regularization, which however effectively reduces model complexity.
Another method is data augmentation, but this depends on understanding the important parameters of data variation.
A third method, semi-supervised learning,  requires a large pool of unlabeled but structurally similar data, which is often not possible in the use-cases cited above.
In this paper we propose a new method, based on the following observation:

A trained NN model that falls short of the fully-trained ``0-1'' ideal may still contain much class-specific information encoded in the responses of the penultimate, pre-softmax layer of neurons. 
This information is not fully utilized by the softmax classifier. 
By using a different final classifier layer that more fully leverages this encoded information, better performance can be extracted from the same NN and the same amount of training data.

\begin{figure*}[t] 
\vskip 0.2in
\begin{center} 
\centerline{\includegraphics[width=\linewidth]{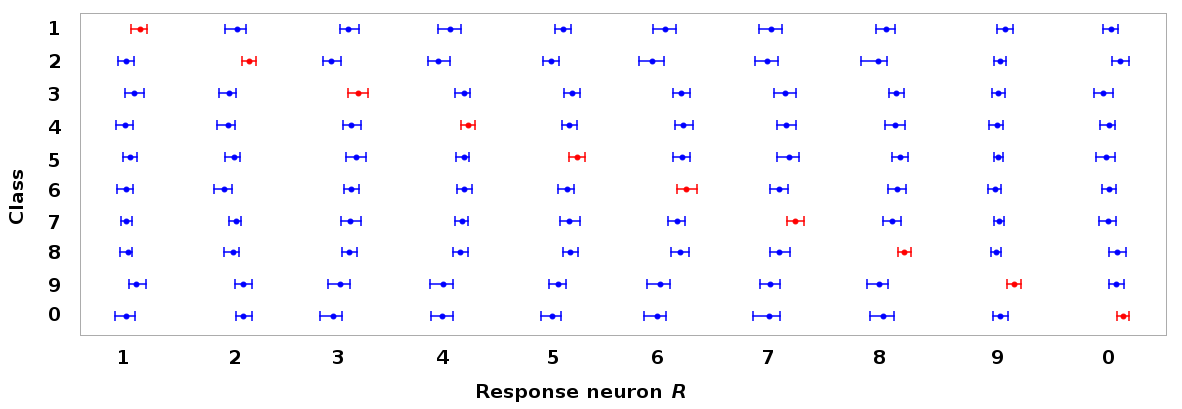}}
  \caption{The full array of class--$R$ response distributions $D$, which  encodes class-specific information in each $R$'s varied responses to the different classes.
 The information in this perspective is not seen by softmax.
 The $j$'th column shows the variety of $R_j$'s responses to the various classes, normalized by the mean of the response to the home class, \textit{i.e.} $\mu_{jj}$.
 Entries are mean $\pm$ std dev, and each column is aligned to show the differences in an $R_j$'s mean responses
 From a Cifar 10 model.   
  }
  \label{classRnDistributionsByCol}
\end{center}
\end{figure*}

The class-specific information considered here is encoded in the responses of the pre-softmax layer neurons.
Each of these neurons (hereafter Response neuron, $R$) develops a characteristic response to each class.
Let $K$ be the number of classes. 
These class - $R$ response distributions form a $K \times K$ array $D$ of probability distributions, where $D_{ij}$ is the distribution of the responses of the $j^{th}$ $R$ to the $i^{th}$ class.
An example of these distributions is seen in Figure \ref{classRnDistributionsByCol}.


Softmax systematically under-utilizes this array $D$.
For a given test sample of true class $i$, softmax uses only the $i$'th row in this array, by comparing the various $R_j$'s responses to samples of class $i$. 
If the NN has not been trained sufficiently to have encoded class separations for class $i$ in this row, softmax will make mistakes.
The core of our proposed approach is to scan the entire array $D$ for clues to the test sample's class. 

Softmax, \textit{e.g.}  \cite{murphyML}, is calculated using the values of the Response neurons $R$ (the penultimate layer) to  sample $s$:
\begin{equation}\label{softmaxEqn}
\hat{s} = \underset{j \in J}{\text{arg max}} \left\{ \frac{e^{R_j(s)} }{\sum\limits_{i \in J}e^{R_i(s)} } \right\} \text{, where }
\end{equation}
$~~~~~~$$\hat{s}$ = predicted class of sample $s$   \newline
$~~~~~~$$R_i(s)$ = response of the $i$th R to $s$  \newline
$~~~~~~$$j \in J$ are the classes (0-9).   \newline

\begin{figure*}[ht!]
\vskip 0.2in
\begin{center} 
\centerline{\includegraphics[width = 1\linewidth]{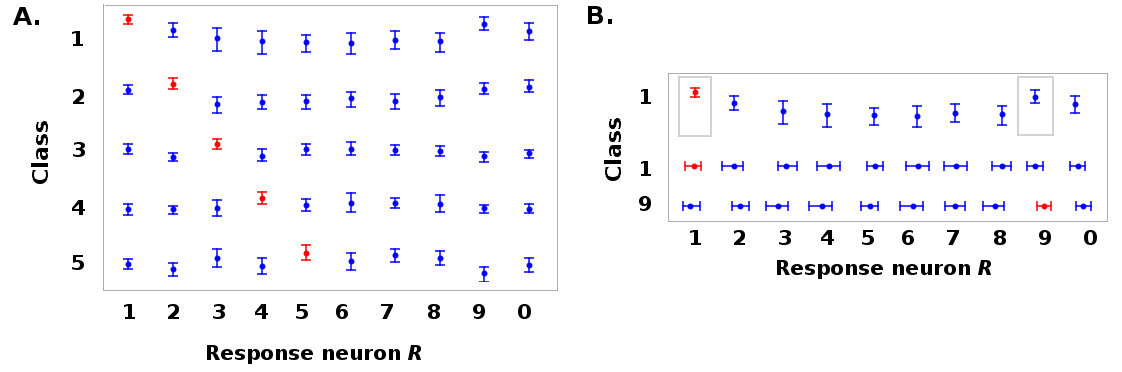}}
  \caption{A: The array of class--$R$ response distributions $D$, as seen by softmax (5 classes shown). 
 The $i$'th row shows the different $R$'s responses to the $i$'th class, normalized by the mean of the response of the home $R_i$, $\mu_{ii}$.
 Entries are mean $\pm$ std dev. 
For a sample with true class $i$, softmax uses only the $i$'th row. 
 B: An example of how extra information can be gleaned from the array $D$.
  Row 1:  Responses of the $R$s to class $1$, normalized by the mean response of the home $R_i$, $\mu_{11}$ (this is row 1 of subplot A).  
Softmax uses only this row to assess a sample from class 1. There is confusion between classes 1 and 9 (in small grey boxes).
 Rows 2 and 3: The responses of the $R$s to class 1 (row 2) and class 9 (row 3), normalized by the mean of $R$'s home class response, \textit{i.e.} by $\mu_{jj}$ for $R_j$,as in Fig \ref{classRnDistributionsByCol} (these are in fact the rows 1 and 9  from Fig \ref{classRnDistributionsByCol}).
The two classes have clearly distinct behaviors in many $R$s, \textit{e.g.} indices  2, 3, 4, 8, and 9.
 From a Cifar 10 model.  
  }
  \label{classRnDistsByRow_useOfDists}
\end{center}
\end{figure*}

During training, a softmax-based optimizer will seek, for the $i$'th class, to maximize the distance  between the response of the $i$'th neuron ($R_i$) to the $i$'th class ($D_{ii}$) and the responses of the other $R$s to class $i$ ($D_{ij},  j\neq i$), by boosting the mean of $D_{ii}$ and depressing the means of $D_{ij}, j \neq i$, and reducing the standard deviations $\sigma_i$ for all $i$.
``Distance'' between two distributions can be described by  the Fisher linear discriminant of two distributions X and Y: 
\begin{equation}\label{fisherDiscriminantEqn}
F(X, Y) = \frac{| \mu_X - \mu_Y |}{0.5(\sigma_X + \sigma_Y)}, 
\end{equation}
\textit{i.e.} the distance between distribution means normalized by their std devs.
When training is ``sufficient'' for softmax, these distances $\{B(D_{ii}, D_{ij}),  j \neq i \}$ become large.
Then, given a test sample with class $i$, softmax yields a very high value in the $i$'th readout, which distinguishes the correct ($i$'th) class using only the $i$'th row of $D$. 
But if training is not ``sufficient'', then $B(D_{ii}, D_{ij})$  will be small for some $j \neq i$. 
In this case, responses will be relatively strong in both $R_i$ and $R_j$, giving a low softmax score and possibly an incorrect estimated class ($j$ instead of $i$).
An example of such confusion, between classes 1 and 9, is seen in row 1 of Figure \ref{classRnDistsByRow_useOfDists}.

However, given uncertainty between classes $i$ and $j$, we can examine the responses of all the $Rs$, relative to the expected statistical behaviors of both the $i$'th and $j$'th classes, shown in Figure \ref{classRnDistsByRow_useOfDists}B.
That is, we can use more than one row of the array $D$ of class--$R$ response distributions, and choose between class $i$ and $j$ by assessing the likelihoods of the $R$ responses given rows $i$ and $j$ of $D$.
If there is confusion between $m$ class responses in the $i^{th}$ row, we can examine $m$ rows.


%

In a case observed by  \cite{delahuntMnist} involving a NN trained by Hebbian updates (\textit{i.e.} not trained via softmax-based loss function), a primitive ``full $D$'' classifier was sometimes much more accurate than softmax (and sometimes it was worse). 
In this paper we describe a Softmax/Pooled-likelihood Hybrid classifier (hereafter \textit{SPH}) that (partially) replaces the softmax layer at testing, in any model for which a validation set is available.
We apply this ``full $D$'' classifier to NN models trained on two datasets: vectorized MNIST and Cifar images.
Our experiments suggest that a trained NN model is in some (not all) cases far more capable than a final softmax layer makes evident, so full use of the response distribution array $D$ can improve test set classification on more difficult test samples.

In our experiments with vectorized MNIST, \textit{SPH} delivered substantial gains in accuracy, using the same trained models, compared to standard softmax.
\textit{SPH}  also outperformed two alternate methods, each built on the same base NN architecture:  an extra fully-connected layer inserted before softmax; and  a Na{\"i}ve Bayes \cite{bayesNotStupid,ngJordanBayes} classifier that replaced the softmax classifier.
This gain in accuracy indicates that the class--$R$ response distribution array $D$ can contain a wealth of information untapped by softmax.
Conversely, it indicates that a ``sufficient'' amount of training data, as measured by softmax performance, actually represents an excess, the surplus training data being required to make up for that part of the encoded class information left on the table by softmax.

The core message of this paper is to highlight the valuable, but currently under-utilized, information contained in $D$. 
A NN whose final classifier layer leverages this extra information encoded in $D$ may be trained on less data yet have equivalent performance to an NN trained on more data but that uses a softmax final layer.
This can potentially ease the training data bottleneck so common in NNs, while still allowing the classifier to hit a given task's performance specs.
The rest of the paper is organized as follows: 
\textit{(i)} Results of \textit{SPH} experiments on a vectorized MNIST dataset and on Cifar images; \textit{(ii)} an overview of the \textit{SPH} algorithm; \textit{(iii)}  discussion.
 

%

\section{Experiments and results}

We ran experiments on vectorized MNIST (hereafter ``\textit{v}MNIST'' to make the vectorization constraint explicit), comparing \textit{SPH} to  standard softmax.
As comparisons we tested two alternative methods of leveraging $R$ on \textit{v}MNIST models: an extra fully-connected layer plus softmax, and a pure Na{\"i}ve Bayes classifier using $R$ as features. 
We also ran experiments on the Cifar datasets (Cifar 10, Cifar 20, and Cifar 100 \cite{lecunMnist,cifar}), comparing \textit{SPH} to softmax.
Results of \textit{v}MNIST experiments indicate the benefits of the \textit{SPH} method, while results of Cifar experiments show its limitations.

\subsection{Setup and results on vMNIST}

Our goal was to see effects of \textit{SPH} vs softmax over a range of training data loads and baseline model accuracies.
We used the \textit{v}MNIST dataset, and controlled trained model accuracy by varying the number of training samples from 100 to 50,000.
We used a simple NN (in Keras/Tensorflow) with two dense layers.
The trained models had mean softmax accuracies ranging from 71\% (given 100 training samples) to 98\% (given 50k training samples).
We then compared the test set accuracies of \textit{SPH}, softmax, and the two comparison methods across this range.   

For each choice of number of training samples we trained and ran 9 models, each with random draws for Train, Test, and Validation sets.
We used 4000 validation and 4000 test samples in all cases.
For each model, we \textit{(i)} trained the model with a softmax-based loss function, as usual; \textit{(ii)} randomly chose non-overlapping validation and test sets from the Test data; \textit{(iii)} ran a parameter sweep over \textit{SPH} parameters using the validation set as described in the Appendix; and \textit{(iv)} recorded test set accuracies using the resulting calculated parameter sets. 

As two Figures-of-Merit we measured raw percentage gain in Test set accuracy  and relative reduction in Test set error  (as a percentage), due to \textit{SPH} versus the softmax baseline.
The second metric allows easier comparison of results on models with widely different softmax accuracies.
For these experiments, we reported the optimal Test set results, \textit{i.e.} we did not use cross-validation on the validation set to fix hyperparameters before proceeding to the Test set.
That is, we set aside the issue of choosing exactly optimal hyperparameters, in order to see what gains were possible (see Appendix for details).

\subsubsection{Gains of SPH vs softmax baseline}
\begin{figure*}[ht!]
\begin{center}
\centerline{\includegraphics[width=0.9\linewidth]{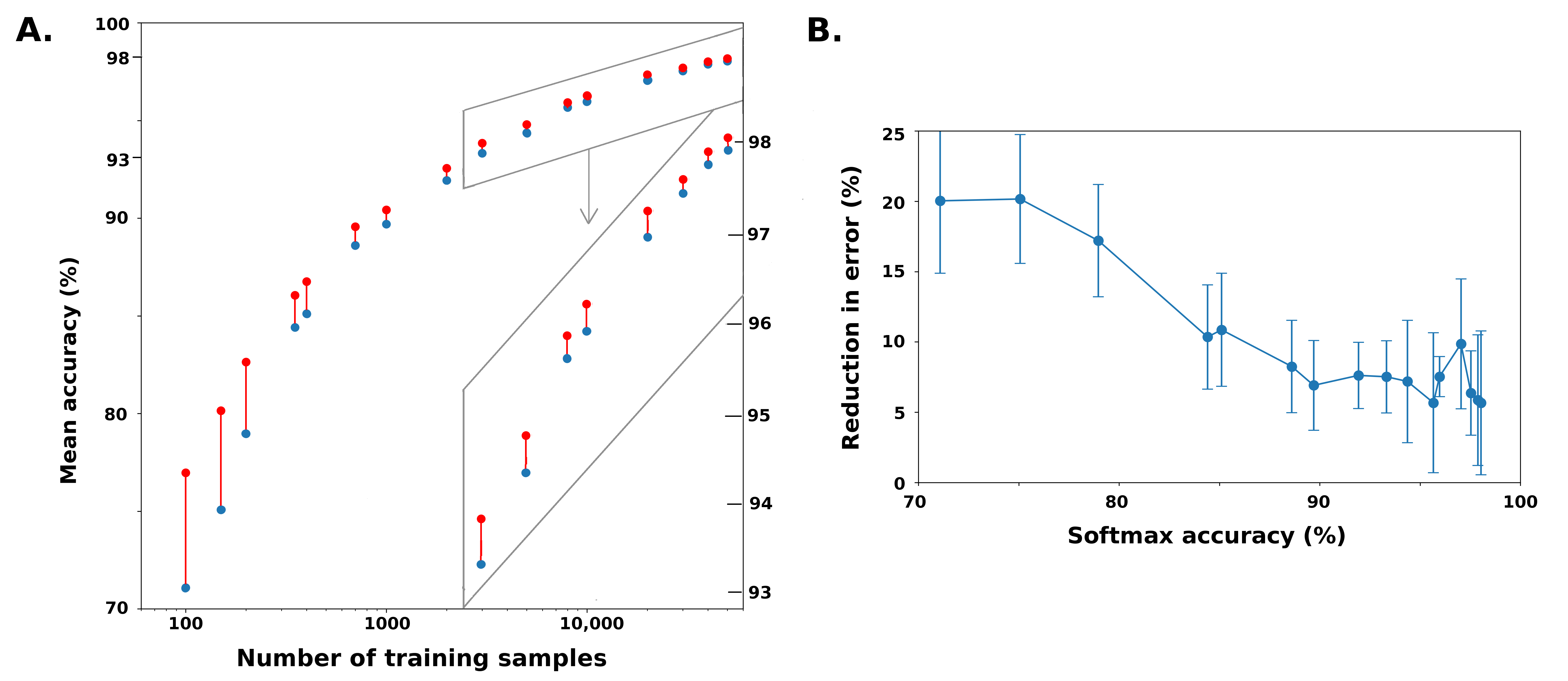}}
  \caption{A: \textit{SPH} gains on \textit{v}MNIST vs number of training samples: 
 Raw percentage gains in accuracy due to \textit{SPH} are shown as vertical red bars connecting softmax baseline (blue dots) and \textit{SPH} accuracy (red dots), for  models  trained on \textit{v}MNIST with varying numbers of training samples (shown on $x$-axis).
 The inset (inside grey trapezoid) shows  models that had softmax accuracy $>$ 93\%, with $x$-axis held fixed and $y$-axis stretched to better show the gains due to \textit{SPH}. 
 B: Relative reduction in Test set error due to \textit{SPH} (\textit{i.e.} $\frac{Err(soft) - Err(SPH) }{Err(soft) } $
 ), vs softmax accuracy, for \textit{v}MNIST models.
  Mean $\pm$ std dev, 9 models generated per number of training samples.
 Replacing softmax with \textit{SPH} reduced Test set error by on average 6 - 20\% for all \textit{v}MNIST models tried. 
  }
  \label{sphGainVsNumTraining}
\end{center}
\end{figure*}

Our core finding is that all  \textit{v}MNIST models benefited from \textit{SPH} at test time, even when the baseline model already had high accuracy.
Test set error was reduced by 6\% to 20\%, with higher reduction in models that were trained on fewer samples (equivalently, models with lower baseline softmax accuracy).
The mean raw percent gains in accuracy are shown as vertical red bars in Figure \ref{sphGainVsNumTraining}A. 
As baseline (softmax) accuracy increased, raw gains from \textit{SPH} over softmax decreased, but the relative reduction in error remained stable even at very high baseline accuracies.
See Figure \ref{sphGainVsNumTraining}B.
 
The gains from \textit{SPH} are virtually free, since it is bolted on after the model is trained as usual.
They result from accessing information encoded in the trained model but unseen by softmax.

\subsubsection{Effects of comparison methods}
To assess the effect of  inserting a fully-connected layer between $R$ and the softmax layer as a way to leverage the information in $R$, we trained a NN with inserted layer on the same training data. 
The NNs with inserted layer showed no gains relative to the simple softmax NN.

We also tested the effectiveness of  replacing the softmax layer with a standard Na{\"i}ve Bayes classifier.
This hybrid NN-Na{\"i}ve Bayes classifier performed consistently worse than the softmax NN.

The failures of these two comparison methods  suggest why \textit{SPH} may appear somewhat complex (even ``hacky''): To successfully extract actionable information from $D$ requires some algorithmic effort.
The various heuristic details of the \textit{SPH} algorithm are what enable it to extract value from $D$.

  
\subsection{Gains from  \textit{SPH}  measured as reduced training data loads}
NNs (especially DNNs) typically require large amounts of training data, which can be time-consuming and costly to collect, annotate, and manage.
In some situations (\textit{e.g.} medical work, field tests, and scientific experiments) data is not only expensive to collect but also constrained in quantity due to real-world exigencies.
In this context, tools that reduce the training data required to hit a given performance spec are valuable.

\begin{figure}[ht!]
\begin{center}
\centerline{\includegraphics[width=0.5\textwidth]{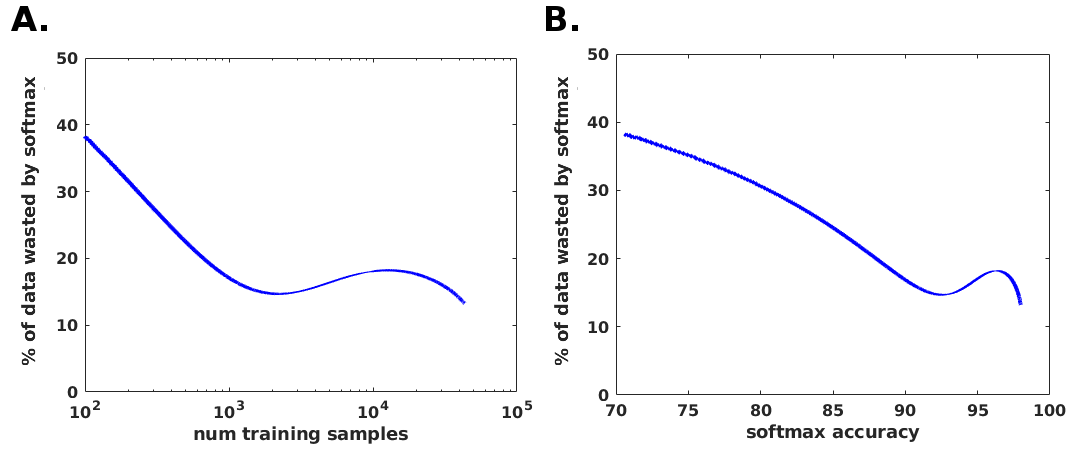}}
\caption{Percentage of training samples wasted by softmax vs \textit{SPH}.
Because softmax does not fully use the class--$R$ distribution matrix $D$, it requires 15\% to 40\% more training samples to attain equivalent Test set accuracy to \textit{SPH}. 
The extra required samples can be thought of as waste due to using softmax.
A: Training data wasted vs number of training samples used by softmax.
B: Training data wasted vs softmax accuracy. 
These curves were made by fitting quadratics to the curves in Figure \ref{sphGainVsNumTraining}, then comparing  training samples numbers required for equivalent Test set accuracies.
  }
  \label{waste}
\end{center}
\end{figure}

\textit{SPH} increased the Test set accuracy of a given model on \textit{v}MNIST, allowing it to match the performance of another model, trained with more data but using softmax. 
Thus, the gain from using \textit{SPH} can be measured in ``virtual training samples'', \textit{i.e.} the number of extra samples that would be needed to attain equivalent test accuracy using only softmax. 
\textit{SPH} yielded a gain of between roughly 1.2x to 1.6x  on \textit{v}MNIST.
That is,  using softmax alone required 15\% to 40\% more training data to attain equivalent accuracy to \textit{SPH}. 
This is plotted in Figure \ref{waste} as ``wasted'' training samples.
Thus, use of \textit{SPH} directly cut training data costs.

  
\subsection{Results on Cifar}

To see whether the \textit{SPH} method worked with deeper NN architectures, we applied it to the Cifar dataset with 10, 20, and 100 classes.
In each case we tested 3 or 4  models of varied trained accuracy (Cifar 10: 59 to 89\%; Cifar 20: 51  to 71\%; Cifar 100: 40 to 60\%).
Models were built from templates at \cite{kerasCifarExamples,MLblogCifar}. 

We found that on Cifar \textit{SPH} yielded only small benefits, or sometimes none at all.
On Cifar 10, relative reductions in Test error were 1.0\% to 2.6\%. 
On Cifar 20, reductions were 0.0\% to 1.4\%. 
On Cifar 100,  reductions were 0.0\% to 0.13\%. 

We see three trends here.
First, the DNNs were much less responsive to \textit{SPH} than the shallow NNs used on \textit{v}MNIST.
Second, datasets with larger numbers of classes were less responsive.
Third, there was some (small) benefit to this approach. 
Whether this indicates that better algorithms than \textit{SPH} might yield useful benefits, or whether DNNs are intrinsically not amenable to this approach, is unknown.



\section{Overview of the \textit{SPH}  algorithm}

\begin{figure*}[h!]
\begin{center}
\centerline{\includegraphics[width=0.8\textwidth]{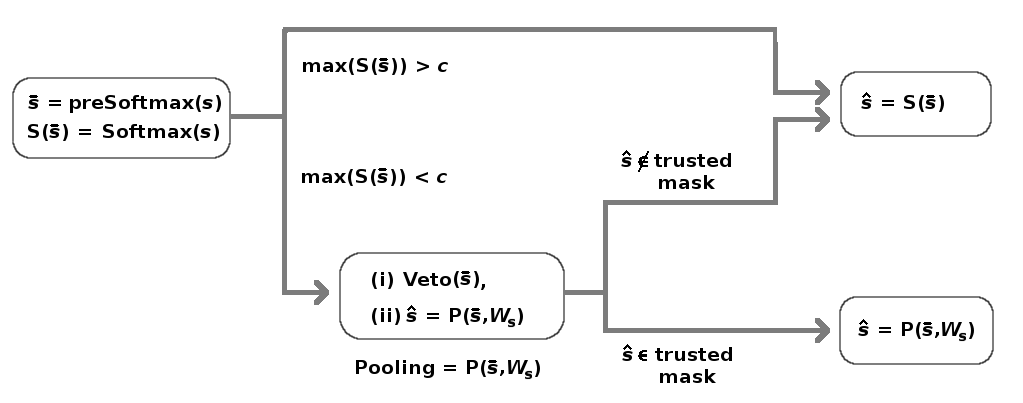}}
  \caption{Flow chart of the \textit{SPH} classifier.
  Given sample $s$, let pre-softmax layer readouts $R(s) = \bar{s}$ and let $\bm{S}(s)$ = the vector of softmax scores for $s$: If the largest softmax score $S(\bar{s})$ is above threshold $c$, we trust the softmax prediction. 
Otherwise we send the sample to the Pooling classifier.
This has two stages, Veto (which vetoes highly unlikely classes), and the pooled likelihood classifier.
If Pooling returns a trusted class (i.e. we trust Pooling to do better than softmax on this class), we keep the Pooling prediction. Else we revert to the softmax prediction.   
  }
  \label{flowChart}
\end{center}
\end{figure*}

This section gives an abbreviated overview of \textit{SPH}.
Full algorithm details are given in the Appendix and in the online codebase \cite{delahuntMothMoneyOnTableCodebase}, which includes Python/Keras code for the \textit{SPH} classifier and for a hyperparameter sweep.
A flow chart is given in Figure \ref{flowChart}. 


We first note that \textit{SPH} is somewhat ``heuristic''. 
As mentioned earlier, the complexities arise because information  from the array $D$ of class--$R$ response distributions does not readily pop out. 
Our intent is not to provide a final algorithm, but to demonstrate that the array $D$ can be usefully mined to improve NN accuracy beyond the abilities of the softmax classifier.

\textit{SPH} combines softmax with a Pooling classifier, which is a species of Na{\"i}ve Bayes  log-likelihood classifier with the $R_j$ responses as input features and a prior on the response distributions $D_{ij}$ which defines them as asymmetrical Gaussians.   
To build the Pooling classifier, we first characterize the  array $D$, by calculating the parameters of each response distribution $D_{ij}$  using a validation set.
We also generate a weight matrix $W$ to capture how well each $R_j$  separates the classes.  
Entries of $W$ are based on the Fisher linear discrimant (see eqn. \ref{fisherDiscriminantEqn}).  

Besides leveraging the diversity of class--$R$ responses of the trained NN, \textit{SPH} takes advantage of two kinds of diversity in sample responses.
First, it distinguishes between samples with high vs low softmax certainty. 
Second, for samples with low softmax certainty it distinguishes  those that are amenable to a Na{\"i}ve Bayes approach from those for which softmax is a more reliable predictor.
Parameters to control  these decisions (trusted softmax scores cutoff, and a list of trusted class predictions) are calculated on a validation set.
 
Suppose a test sample $s$ is passed into the model,  and let $\bm{S}(s)$ = the vector of softmax scores for $s$. 
If the top softmax score $max(\bm{S}(s))$ is high then we trust the softmax prediction, since a high softmax score typically indicates that the softmax estimate is correct, and we are done.
If the top softmax score is low, we route the sample to the Pooling classifier, which uses the information in $D$ to make a prediction.  
If the prediction returned by the Pooling classifier is among the set of ``trusted'' classes then we trust the Pooling prediction. 
If it is in the set of ``untrusted'' classes, then we revert to the softmax prediction.
Thus, \textit{SPH} takes advantage of the fact that softmax and Pooling each work best on different types of samples.  

\paragraph{Training}
The model is trained as usual. 
We note that training with softmax is not required: \textit{SPH} can be used with any model, regardless of optimization method, that has a layer of  $K$ response units $R_j$s.
Here we only examined softmax-optimized NNs, so the $R_j$s are the $K$ pre-softmax units, with $R_j$ corresponding to the unit that targets the $j$th class.

\paragraph{Defining the response distribution array D}
As internal NN weights update during training, each $R_j$ develops distinct responses to each of the $K$ classes. 
Let $D_{ij}$ equal the response distribution of $R_j$ to samples of class $i$. 
To characterize $D_{ij}$, we first run a Validation set through the trained NN.
Of these validation samples, we select only those within a certain range of softmax scores (high and low limits are hyperparameters) to characterize $D$.
An optimal range will likely not include high-scoring samples, because the distributions in $D$ are only relevant to samples that will be classified by Pooling.
These sample are low-scoring by design, so we wish the array $D$ to target this population.

We define each $D_{ij}$ as an asymmetrical Gaussian, with a mean  (or median if wished) $\mu_{ij}$ and separate left and right standard deviations (\textit{i.e.} $\sigma_{ij}^L \text{ and } \sigma_{ij}^R$).  
This characterizes $D$ as  three $K \times K$ arrays: 
\begin{equation} \label{defnOfDEqn}
D_{ij} := \{  \mu_{ij}, \sigma^L_{ij}, \sigma^R_{ij} \} 
\end{equation}
%

\paragraph{Weight matrix W}
Not all $D_{ij}$ are created equal. 
In some cases,  $D_{ij}$ contains valuable class-separating information, while in other cases, an $R_j$'s responses to different classes overlap, and $D_{ij}$ is noisy for some (not all) $i$'s.
We encode  the usefulness of $D_{ij}$s in a weight matrix $W$ (size $K \times K$), to weight the likelihoods of the various classes at testing. 
%
We wish a function $w_{ij}$ that reflects whether $R_j$ distinguishes class $i$ from all the other classes.
We define 
\begin{equation}
w_{ij} = \underset{k \neq i}{\mathrm{median}} \{  \frac{ ( \mu_{ij} - \mu_{kj} ) }{0.5 (  \sigma^L_{ij} +  \sigma^R_{kj} ) } \},
\end{equation}
(assuming $\mu_{ij} > \mu_{kj}$. A similar formulation with $\sigma^L$ and $\sigma^R$ reversed works for the opposite case). 
This is done for each $i,j$ because a given $R_j$ may separate some class $i$'s from the other classes well, and some class $i$'s badly. 
We note that this is a compromise solution which loses some class-specific information. 
$W$ is then sparsified by setting low $w_{ij} = 0$ since these indicate non-separating $D_{ij}$.
Then $W$ is normalized by row (\textit{i.e.} for each class, over all $R$s) because for a test sample $s$,  the vector of pre-softmax NN outputs $R(s) = \bm{s}$ is what will be visible.

\paragraph{Class mask $\bm{m}$}
Classes display diverse behaviors when passed through a trained NN, and they will each benefit more or less from a Pooled-likelihood classifier $P()$ versus softmax $S()$.
For a sample $s$,  let $P(s)$ and $S(s)$ be the predicted classes via Pooling and via softmax. 
If a class $i$ is poorly classified by Pooling, we wish to distrust and ignore the prediction $\hat{s} = P(s) \text{ whenever } \hat{s} = \text{class }i$, and revert to the softmax prediction $S(s)$ instead.
We encode this choice in a logical $1 \times K$ vector $\bm{m}$.

To estimate which classes respond well to Pooling, we use the samples in a Validation set that have low softmax scores (\textit{i.e.} $\{ s: max(\bm{S}(s) )< c\}$ where $c$ is a hyperparameter).
We run these {$s$} through the Pooling classifier, and calculate the accuracy on each class. 
Let $V_i$ be the set of low-scoring validation samples from class $i$. 
If $\text{accuracy}(P(V_i) ) -  \text{accuracy}(S(V_i) )  >   a$  (\textit{e.g.} $a = 0$)
then we set $\bm{m}_i = 1$, to indicate that Pooling will (hopefully) give better results than softmax on these low-scoring samples at testing;  
else $\bm{m}_i = 0$ and  we should not trust Pooling if $P(s) = i$. 

\paragraph{Testing}
At test time, a sample $s$ is run through the NN as usual.
If its softmax score is high ($max(\bm{S}(s) )< cc$) the softmax prediction is accepted.
If the softmax score is low, the sample is sent to the Pooling classifier.

\paragraph{The Pooling classifier}
The Pooling classifier has two stages: a veto stage, and a log-likelihood predictor. 
In the veto stage, the vector of the $R$s responses to sample $s$ ($R_j(s) = \bm{s}_j)$ are compared to the distributions $D$. 
If $\bm{s}_j$ falls far outside the expected behavior of class $i$, for enough $R_j$, then class $i$ is vetoed, \textit{i.e.} it is removed from consideration as a possibility for $\hat{s}$.
In the predictor stage, the pooled likelihood measure is applied using a sample-specific weight matrix $W(s)$.
If the predicted class $\hat{s} = P(s)$ is trustworthy (according to $\bm{m}$), we keep this Pooled prediction; else we revert to the softmax prediction.

Thus, \textit{SPH} uses the Pooled classifier on harder (more uncertain) samples, but only when Pooling predicts a class for which we expect Pooling to succeed, based on results on the validation set.
\textit{SPH} uses the softmax prediction on easier samples and also on  samples  where Pooling predicts a class for which we expect it to do poorly.



%
\section{Discussion}

Neural Nets are more capable than we realize, and contain reserves of class-specific information untapped by softmax.
This is particularly true when one lacks abundant training data.
But limited data, even scarcity of data,  is the norm in many important ML use cases, including medicine, scientific experiments, and field-collected data. 
In this event we wish to maximize our use of the class-specific information that training has encoded in the model.
Softmax may be a sub-optimal way to do this in some cases.

We consider a situation where two constraints, problem complexity and limited data, clash.
Model complexity is driven by the problem's complexity, while limited data is a separate axis of constraint. 
Suppose a problem requires for its capture a model of complexity C, but there is not enough data to properly train that model. 
One option is to use a more parsimonious model, or regularization. 
But these methods reduce model complexity below that required by the problem.
Data augmentation is sometimes another option, but this is not always technically possible. 
In this paper we suggest a third way to address the mismatch between problem complexity and available data, namely, ``Don’t throw away information''.
 
In particular, the hybrid classifier \textit{SPH}  combines softmax and a Na{\"i}ve Bayes-like pooled-likelihood method, in order to leverage class-specific behaviors encoded in the trained model which are ignored by softmax. 
%

\textit{SPH} focuses on two  forms of inter-class diversity:
\textit{(i)} The class--$R$ response distributions (where $R$s are the pre-softmax neurons) contain a wealth of class-specific information, encoded by the NN during training, which can be tapped to improve model performance. 
We leverage this information by defining a pooled likelihood classifier over the array $D$ of the class--$R$ response distributions.
\textit{(ii)} Classes display diverse behaviors as they pass through the model. 
In particular, softmax may tend to fail on certain classes more than others.
We leverage this diversity by defining a mask that determines which Pooling predictions are trustworthy, and which should be ignored, trusting instead on the softmax predictions.

We note there is nothing at all magical or optimal about \textit{SPH} as a means to leverage information encoded in $D$.  
While the heuristic complexities of \textit{SPH} are perhaps justified by the failure of simpler comparison methods (\textit{viz.} an inserted fully-connected layer, and a Na{\"i}ve Bayes  classifier), we are fairly certain that more effective approaches to this problem exist.
Possible avenues include: 
\textit{(i)} find algorithms that better utilize $D$; 
\textit{(ii)} dig  into the NN's inner layers, to find salient class-specific behaviors; and 
\textit{(iii)} during training, use loss functions that directly utilize $D$ (instead of softmax), in order to accentuate class-specific contrasts in the final trained $D$.

In our experiments on NNs trained with softmax-based loss functions, \textit{SPH} delivered significant gains for shallow NNs trained on the \textit{v}MNIST dataset, but minimal gains for DNNs trained on Cifar images. 
These clear gains for shallow NN models indicate that for some uses cases and datasets,  money can be made by replacing softmax with an alternative such as \textit{SPH}. 

However, \textit{SPH} failed to benefit DNNs.
 We suspect that one reason was that trained softmax scores tended to bunch near 1.0 even when the model was  wrong. 
This impacted the effectiveness of the cutoff $c$ (used to decide when to use Pooling vs softmax):
On one hand,  $c$ does not need softmax scores to be well-calibrated  to actual probabilities (fortunately,  since many DNNs are poorly-calibrated \cite{guoCalibration}).
It is simply an empirically-determined cutoff, so the scores’ overall relationship to probabilities is not relevant as long as the scores are well ordered. 
But if the scores of similar input samples have large variance, so that samples with similar probabilities have widely varying softmax scores, then a cutoff $c$  becomes an unreliable guide. 
Thus what matters is the precision, not the accuracy, of the calibration. 
 DNNs may have failed to show benefit because if all the softmax scores are squeezed into a small region near 1.0, there is less dynamic range.
Then the effective precision of the softmax scores is lower, making the threshold $c$ less tied to actual model certainties.

The  general approach described here might be potentially useful for any trained model that contains accessible class-response distributions such as the array $D$.
We experimented only on NNs trained with a softmax loss function, which would tend to maximize the softmax-accessible information encoded by training. 
It is an open question whether training a given model (including DNNs) with non-softmax loss functions can make the information encoded in $D$ more accessible to methods such as \textit{SPH}, further improving accuracy over softmax.



%

\section{Appendix: Algorithm details}
This Appendix gives details about the Softmax-Pooling Hybrid (\textit{SPH}) algorithm.
It has two main parts: \textit{(i)} Determining hyperparameters, including $D$; and \textit{(ii)} applying \textit{SPH} to a Test set.
Full code for running the method can be found at  \cite{delahuntMothMoneyOnTableCodebase}.


\subsection{Determining hyperparameters} \label{setHyperparameters}

This section discusses how to prepare resources for \textit{SPH}.
We need the following: \textit{(i)} the array of class-$R$ response distributions $D$; \textit{(ii)} a weight matrix $W$; \textit{(iii)} a class mask $\bm{m}$; \textit{(iv)} assorted other hyperparameters.
$W$ and $\bm{m}$ both depend on hyperparameters, while $D$ does not.


\subsubsection{Characterize $D$, the array of class-$R$ response distributions}
During the training phase, as internal NN weights update, each unit in the pre-softmax layer $R$ develops distinct responses to each of the $K$ classes. 
Let $D_{ij}$ equal the response distribution of $R_j$ to class $i$. 
To characterize $D_{ij}$, we first run a Validation set through the trained NN.
(We cannot use the Training set for this, since samples used to train internal model weights have distinct behavior when passed through the trained model).

Of these validation samples, we select only those with softmax scores within a certain range (high and low limits are hyperparameters) to characterize $D$.
An optimal range will likely exclude high-scoring samples, because the distributions in $D$ are only used on samples  passed to Pooling, which are low-scoring by design. 
We wish $D$ to target this population.

We define $D_{ij}$ as an asymmetrical Gaussian, via a mean (or median if wished) $\mu_{ij}$ and separate left and right standard deviations  $\sigma^L \text{ and } \sigma^R$.
We note this prior is intermediate between a pure Gaussian and a Na{\"i}ve Bayes-style distribution generated purely from validation samples. 

$\mu_{ij}$ is simply the mean (or median) of $R_j(s) | s \in \text{class } i $.
 
We generate $\sigma^R_{ij}$ via mirror images: Let $R$ = the set of responses $R_j(s)~|~ \{s \in \text{class } i \land R_j(s) > \mu_{ij} \}$ where $\land$ = logical AND. Then subtract $\mu_{ij}$, mirror this set, and calculate the std dev:  $\sigma^R_{ij}$ = std dev$(  [ R - \mu_{ij} ], [ - R + \mu_{ij} ] )$. A similar calculation gives $\sigma^L_{ij}$.

Doing this for each class-$R$ pair $i,j$ characterizes $D$ as  three $K \times K$ arrays:
\begin{equation}
D_{ij} = \{  \mu_{ij}, \sigma^L_{ij}, \sigma^R_{ij} \} 
\end{equation}
 

\subsubsection{ Weight matrix $W$}
Not all $D_{ij}$ are created equal. 
In some cases,  $D_{ij}$ contains valuable class-separating information, while in other cases, an $R_j$'s responses to different classes overlap, so that $D_{ij}$ is noisy for some (not all) $i$'s.
We encode these differences in a weight matrix $W$.
When assessing the likelihoods of the various classes at testing, $W$ emphasizes some $R$s over others, different for each class. 
The process has three parts, described below: \textit{(1)} calculate a variant of Fisher distance for each $D_{ij}$; \textit{(2)} set $W_{ij} = 0$ for noisy $D_{ij}$; \textit{(3)} assign positive values to the remaining $W_{ij}$.\\  
 \textit{(1)} We wish a variant of Fisher distance $f_{ij}$ that reflects whether $R_j$ distinguishes class $i$ from the other classes.
Define \\
\begin{equation} \label{fisherDiscriminantEqn}
f_{ij} = \underset{k \neq i}{\mathrm{median}} \{  \frac{ ( \mu_{ij} - \mu_{kj} ) }{0.5 (  \sigma^L_{ij} +  \sigma^R_{kj} ) } \} 
\end{equation} 
(assuming $\mu_{ij} > \mu_{kj}$; a similar formulation  with $\sigma^L$ and $\sigma^R$ reversed works for the opposite case).
This is done for each $i,j$ because a given $R_j$ may separate some class $i$'s from the other classes well, and some class $i$'s badly (\textit{e.g.} in a highly-trained softmax NN, class $i$'s ``home unit'', $R_i$,  is optimized to distinguish class $i$ best of all).
We note that use of \textit{median()} loses some class-specific information, because $f_{ij}$ ignores how $R_j$ separates classes $\{k, p ~| ~k, p \neq i \}$. \\ 
 \textit{(2)} Sparsify $W$: For some $i,j$ pairs the class distributions are just too overlapped. 
We set $w_{ij} = 0 ~ \forall f_{ij} < $ some  threshold $w_1$. For example, if we wish to only consider classes  at least 2 std dev apart, we set $w_1$ = 2.\\ 
\textit{(3)} We set $w_{ij} = f_{ij}~ \forall f_{ij} > w_1$. 
Then normalize the rows of $W$: $w_{ij} = \frac { {w_{ij}}^\alpha } {  \sum_{j} { {w_{ij}} ^\alpha } }$,  where $\alpha$ is a sharpener, and each new $w_{ij}$ is calculated using old values, not the new values.
We normalize by row (\textit{i.e.} for each class, over all $R_j$s) because for a test sample $s$,  the vector of pre-softmax NN outputs $\bm{R}(s) = \bm{s}$ is the visible output.


\subsubsection{Class mask}

Classes display diverse behaviors when passed through a trained NN, and they will benefit (or suffer) more or less from a Pooled-likelihood classifier $P()$ versus softmax $S()$.
For a sample $s$,  let $P(s)$ and $S(s)$ be the predicted classes via Pooling and via softmax. 
If a class $i$ is poorly classified by Pooling, we wish to distrust and ignore the output $P(s) \text{ whenever } P(s) = \text{class }i$, and revert to the softmax prediction instead.

To estimate which classes respond well to Pooling, we use the samples in a Validation set that have low softmax scores (\textit{i.e.} $\{ s: max(\bm{S}(s) )< c\}$ where $c$ is a hyperparameter).
We run these {$s$} through the Pooling classifier, and calculate the accuracy on each class. 
Let $V_i$ be the set of low-scoring validation samples from class $i$. 

If $\text{accuracy}(P(V_i) ) -  \text{accuracy}(S(V_i) )  >   a_1$  (where $a_1$ is a hyperparameter, \textit{e.g.} $a = 0$), we set $\bm{m}_i = 1$, to indicate that we expect Pooling to give better results than softmax on  low-scoring test samples from class $i$; else $\bm{m}_i = 0$ and  we should not trust the Pooling prediction $P(s) = i$.

$\bm{m}$ is used by \textit{SPH} as follows: Suppose the class predicted for a sample $s$ by Pooling $\hat{s} =P(s) = i$, then if $\bm{m}_i = 1$ we trust the result; else we use the softmax result, $\hat{s} = S(s)$.


\subsubsection{Hyperparameters}
These include (organized by purpose):\\ 
1. To determine which samples are routed to Pooling:\\
$~~~~c$ = certainty threshold: For a sample s, if $max(S(\bm{s})) < c$ then $s$ gets re-routed to the pooling classifier. \\ 
2. To determine which samples to use when characterizing $D$, by keeping $max(S(\bm{s}))$  within a relevant range: \\
$~~~~c_l$ = lower threshold (eg $c_l = c - 0.3$);\\
$~~~~c_h$ = upper threshold (eg $c_h = c + 0.2$). \\ 
3. To determine the weight matrix $W$:\\
$~~~~w_1$: minimum Fisher distance threshold, used to cull noisy class-$R$ distributions. A high value makes $W$ sparser.\\ 
$~~~~w_2$: an exponent to sharpen the contrasts in different $w_{ij}$.\\ 
4. To control the Veto stage:\\
$~~~~v_1$: For sample $s$, if class-$R$ mahalanobis distance $M_{ij}(s) > v_1$, then $R_j$ is suggesting that class {i} is improbable. \\
$~~~~v_2$: For sample $s$ and class $i$, if $v_2$ $R_j$s trigger the $v_1$ improbability flag,  then (for $s$ only), class $i$ is veto'ed as a possible prediction.\\ 
5. To determine which classes have trustworthy Pooled results (\textit{i.e.} $\bm{m}$):\\
$~~~~a_1$: defines an expected gain of Pooling over softmax. 
In the Validation set, if for class $i$ the Pooled accuracy does not exceed softmax accuracy by at least $a_1$ (on low-scoring samples), Pooling predictions  $P(s) = i$ are ignored.\\ 

To optimize hyperparameters via a parameter sweep, we \textit{(i)} choose a set (range) of values for each hyperparameter; \textit{(ii)} create $W$ and $\bm{m}$ on the Validation set; then \textit{(iii)} re-apply \textit{SPH} to the Validation set.
This is fast in practice, since the NN model only needs to run on the validation and Test sets once, and the process of assessing hyperparameter sets admits of various shortcuts.

\begin{figure*}[ht!]
\vskip 0.2in
\begin{center}
\fbox{
\rule[0cm]{-0.25cm}{-1cm}
\centerline{\includegraphics[width=\linewidth]{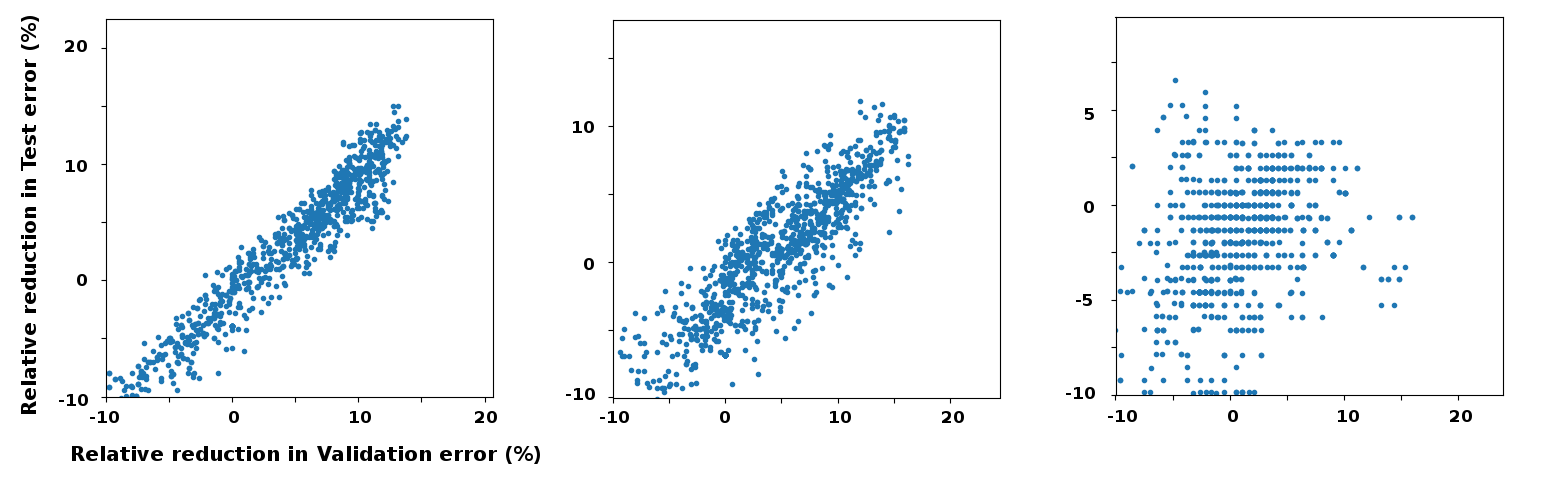}}
}
\caption{Examples of correlation between  reductions in error due to \textit{SPH} on validation ($x$-axis) and Test sets ($y$-axis). 
Each datapoint corresponds to a hyperparameter set.
Results are for three vectorized MNIST models with validation accuracy (left to right) 74\%, 84\%, 95\%. 
In some cases the correlation is close, so selecting hyperparameters based on validation set gains is easy (\textit{e.g.} lefthand plot).
In other cases, it is not obvious how to select hyperparameters, based solely on validation set gains, that will yield optimal test set gains (\textit{e.g.} righthand plot).}
\label{testVsValReductionInError}
\end{center}
\vskip -0.2in
\end{figure*}

In this work we elided the question of how to choose the hyperparameter set based solely on Validation set outcomes.
In some cases there is a clean correlation between gains on validation and test sets.
In other cases, selecting the hyperparameters which give maximum validation gains yield sub-optimal Test set gains.
Figure \ref{testVsValReductionInError} shows a range of scenarios.
We expect that in general some kind of cross-validation is required to optimally select hyperparameters for generalization to Test sets.


\subsection{Applying \textit{SPH} to a Test set}
To apply \textit{SPH} to a test sample $s$, we have the following order of events (see flow chart in Fig \ref{flowChart}): \\
$~~~$\textit{1.} Send $s$ to softmax or to Pooling, based on softmax score.\\
$~~~$\textit{2.}  If $s$ is sent to softmax, $\hat{s} = S(s)$, and we are done. If $s$ is sent to Pooling, there are three steps:\\
$~~~~~$\textit{a.} The veto step removes certain classes from consideration via a new weight matrix $W(s)$.\\ 
$~~~~$\textit{b.} The pooled likelihood measure, using $D$ and $W(s)$, returns a Pooled prediction $P(s) = i$.\\
$~~~~$\textit{c.} The class mask $\bm{m}$ is applied: If $i$ is a trusted class for Pooling, we keep the Pooling prediction, $\hat{s} = P(s)$, and we are done. 
Else we revert to the softmax prediction, $\hat{s} = S(s)$.\\


Below we detail  the various steps:

\subsubsection*{1. Gate samples using softmax scores}

For a sample $s$ let $\bm{S}(s)$ = the vector of  softmax readouts.
Softmax tends to be reliable when its top score $max(\bm{S}(s))$ is high, and unreliable when this  top score is low.

We apply a gating threshold: 
If $max(\bm{S}(s)) \geq c$, we accept the softmax prediction $\hat{s}  = S(s)$ and we are done.
If $max(\bm{S}(s))  < c$, we route $s$ to the Pooling branch.

\subsubsection*{2. Pooling branch}

This branch has three steps.
As preparation we define asymmetric Mahalanobis distance for use in steps \textit{a} and \textit{b}. 
Mahalanobis distance measures how far a sample is from the mean of a distribution, using the standard deviation $\sigma$ as the unit of distance: 
\begin{equation}
M(x) = \frac {|x - \bar{x}| } { \sigma }
\end{equation}
For a Gaussian, $M^2$ is the log likelihood.
The Pooling branch uses an asymmetrical Mahalanobis distance of $\bm{s}_j$ from the centers of  each $D_{ij}$, where standard deviations are different on either side of the center (\textit{median()} may be used instead of \textit{mean()}, to downplay outliers).
Define  the asymmetrical Mahalanobis distance of $\bm{s}_j$ from $D_{ij}$ as follows: 
\begin{multline}
 ~~~~~~~M_{ij}(\bm{s_j}) = \frac{ (\mu_{ij} - \bm{s}_j )}{ \sigma^L_{ij} } ~ ( \text{if } \mu_{ij} > \bm{s}_j), \text{ or} \\
 M_{ij}(\bm{s}) = \frac{ (\bm{s}_j - \mu_{ij} )}{ \sigma^R_{ij} } ~ (\text{if } \mu_{ij} < \bm{s}_j)~~~~~~~~~~~~~~~ 
\end{multline}


\subsubsection*{2a. Veto stage}
We use the distributions array $D$  to rule out ``impossible'' class predictions.
Roughly speaking, if $\bm{s}_j$ is many standard deviations ($ \sigma^L_{ij}$ or $ \sigma^R_{ij} $) from $\mu_{ij}$ for some $j$, then class $i$ is very unlikely to be correct: 
The behavior of sample $s$ does not fit the distributions of class $i$. 
On the other hand, sometimes samples happen to fall in the outer reaches of their class distributions, especially when $N$ is high (eg in Cifar 100). So we do not want to veto a class due to just one outlandish $R$ response. 

We use two parameters, $v_1$ and $v_2$. 
$v_1$ is the number of standard deviations that triggers an outlier status.
$v_2$ is the number of $R$s that must be triggered to veto a class. 
For sample $s$, let $\text{I}_{ij}(s) = 1$ if $M_{ij}(\bm{s}) \geq v_1$, 0 otherwise. 
We wish to veto all classes $i$ for which $ \sum_{j} \text{I}_{ij}(s) \geq v_2$.

We create a new sample-specific weight matrix $W(s)$ from $W$ by  zeroing out the $i$'th row of $W$ ($w_{ij} = 0~ \forall ~j$), for each such class $i$.
This removes the vetoed classes from consideration. 


\subsubsection*{2b. Pooled-likelihood classifier}

We use a weighted asymmetric Mahalanobis distance to predict the class of $s$: 
\begin{equation}
 \hat{s} = P(s) = \underset{classes~ i}{\mathrm{min (>0) }} \{~  \sum_{j} (~{W(s)}_{ij}M_{ij}~)^{m_2}~ \}
 \end{equation}
where $m_2$ is a sharpener.
This is the weighted sum of each row of $M$, a quantity similar to a log likelihood for each class.



 \subsubsection*{2c. Class mask}
Based on the diverse behaviors of classes in the Validation set, we expect Pooling to do relatively better than softmax on some classes and worse on others. 
This is encoded in the masking vector $\bm{m}$, where $\bm{m}_i = 1 $ if Pooling is good at classifying class $i$, and 0 otherwise.
At test time we see only the predicted classes $\{P(s)\}$, not the true classes, so we mask based on predicted class label:
Suppose the Pooling branch prediction $P(s) = i$. If $ \bm{m}_i = 1$, we accept the Pooled prediction, $\hat{s} = P(s)$. 
If $\bm{m}_i = 0$, we ignore the Pooled result and revert to the softmax prediction, $\hat{s} = S(s)$.

\nolinenumbers

\section*{Acknowledgments}
\noindent CBD's work was partially supported by the Swartz Foundation.\newline 
JNK acknowledges support from the Air Force Office of Scientific Research (FA9550-19-1-0011).




%

\bibliography{mothBibliography_june2019}
\bibliographystyle{IEEEtran}

\end{document}